\documentclass[11pt]{article}

\usepackage[margin=1in]{geometry}
\usepackage{amsmath,amssymb,amsthm}
\usepackage{graphicx}
\graphicspath{{figures/}}
\usepackage{booktabs}
\usepackage{multirow}
\usepackage{array}
\usepackage{xcolor}
\usepackage{hyperref}
\usepackage{algorithm}
\usepackage{algpseudocode}
\usepackage{microtype}
\usepackage[numbers,square,sort&compress]{natbib}
\usepackage{caption}
\usepackage{subcaption}
\usepackage{placeins}   
\usepackage{booktabs}
\usepackage{tabularx}

\usepackage{verbatim}

\hypersetup{colorlinks=true, linkcolor=blue, citecolor=blue, urlcolor=blue}

\title{\textbf{DeepPySR - A Symbolic Regression Framework with Dynamic Pruning, Pareto Selection, and Hierarchical Composition for Real-World Scientific Discovery}}

\author{%
  \parbox{0.92\textwidth}{\centering
  Fuling Chen$^{1,*}$,\; Kevin Vinsen$^{1}$,\; Phillip Melton$^{2,3}$,\; Rae-Chi Huang$^{4}$ \\[6pt]
  \small
    $^1$International Centre for Radio Astronomy Research (ICRAR), University of Western Australia, Perth, WA, Australia \\ $^2$The Raine Study, University of Western Australia, Perth, WA, Australia \\ $^3$Menzies Institute for Medical Research, University of Tasmania, Hobart, TAS, Australia \\ $^4$Nutrition \& Health Innovation Research Institute, School of Medical and Health Sciences, Edith Cowan University, Perth, WA, Australia \\[4pt] $^*$Correspondence: \texttt{fuling.chen@uwa.edu.au}
  }%
}

\date{}

\begin{document}
  \maketitle

  \begin{abstract}

Symbolic regression (SR) discovers analytical equations from data, yielding glass-box models with directly interpretable formulas, unlike black-box methods that rely on unstable post-hoc tools such as SHAP or LIME. 
This transparency is crucial in clinical medicine and social science, but SR faces three challenges: high-dimensional inputs, principled selection of Pareto-front formulae, and data irregularities such as multicollinearity and class imbalance. 
We introduce DeepPySR, which addresses these issues with a dynamic variable-pruning schedule to remove irrelevant features during search, an exponential Pareto selection criterion that eliminates trade-offs between accuracy and complexity, and a multi-layer architecture for hierarchical symbolic composition. 
On four Feynman physics benchmarks and seven biomedical and social-science datasets, DeepPySR outperforms PySR and baselines on body fat (R$^2$: 0.794 vs.\ 0.702), heart disease (F1: 0.898 vs.\ 0.787), student performance (R$^2$: 0.964 vs.\ 0.948), and Raine BMI (R$^2$: 0.525 vs.\ 0.370), producing interpretable formulas aligned with domain risk factors.
  \end{abstract}
  \section{Main}

  The tension between predictive accuracy and interpretability is a central problem in applied machine learning.  In scientific disciplines, from biomedicine to astrophysics, understanding \emph{why} a model makes a prediction is often as important as the prediction itself.  This need is especially acute in \emph{high-stakes} domains such as clinical medicine, public health, and social science, where model decisions directly affect individuals and must be auditable, legally defensible, and comprehensible to domain experts who are not machine-learning practitioners~\cite{rudin2019stop}.  A cardiologist does not need just a risk score but the mechanistic factors driving it; an epidemiologist studying childhood BMI trajectories needs an equation that can be communicated and acted on; an astronomer needs more than a candidate ranking but the physical process that caused the ranking.

  The prevailing response has been to train high-accuracy black-box models and apply \emph{post-hoc} explanation tools such as SHAP~\cite{lundberg2017unified} or LIME~\cite{ribeiro2016why} to approximate local feature attributions.  This strategy has well-documented shortcomings: post-hoc explanations are approximations of a separate model, are \emph{unstable} under small perturbations~\cite{alvarez2018robustness}, and can be \emph{manipulated}~\cite{slack2020fooling}.  No amount of attribution substitutes for knowing the actual function the model computes~\cite{rudin2019stop}.

  Symbolic regression (SR) offers a different paradigm: it directly searches for an analytical, human-readable mathematical expression that fits the data~\cite{koza1994genetic,cranmer2023interpretable}.  The field has seen renewed interest following the release of PySR~\cite{cranmer2023interpretable} and the Feynman SR benchmark~\cite{udrescu2020ai}.  Kolmogorov-Arnold Networks (KAN)~\cite{liu2024kan} have been proposed as a gradient-based interpretable alternative, but KAN requires a pre-specified architecture and pre-selected features~\cite{chen2026longitudinal}, making it impractical when the relevant feature set is unknown.

  Real-world biomedical datasets impose additional difficulties.  \emph{Intraclass correlation} (ICC) is ubiquitous: weight, waist circumference, and BMI z-scores are near-collinear, and standard SR collapses onto correlated variables without improving generalisation.  \emph{High dimensionality} ($d \gg 10$) overwhelms evolutionary search.  \emph{Class imbalance} (e.g.\ stroke events, $<$5\% prevalence) degrades classification when handled only by resampling.  \emph{Formula selection} from the Pareto front remains heuristic, and \emph{hierarchical structure} cannot be captured by a single-layer $y = f(\mathbf{x})$ formulation.  Prior work has addressed feature selection as a pre-processing step~\cite{hein2018interpretable,virgolin2020improving} and explored multi-layer SR~\cite{kim2020integration}, but no framework integrates these components in a modern evolutionary engine.

  We present \textbf{DeepPySR}, an extension of SymbolicRegression.jl~\cite{cranmer2023interpretable}, with three contributions.  
  \begin{enumerate}
      \item ~\textbf{Dynamic Variable Pruning Schedule (DVPS)}: a scheduled mutation operator that linearly ramps up pruning pressure during evolutionary search, automatically eliminating irrelevant features without a separate pre-processing step, and making the framework usable on datasets with hundreds of variables.  
      \item ~\textbf{Exponential Pareto Selection (EPS)}: a principled, user-tunable scoring function that selects the single best formula from the Pareto front by jointly rewarding accuracy and penalising complexity exponentially; the parameter grid search is computationally negligible, as it scores existing hall-of-fame equations rather than retraining the model.
      \item ~\textbf{Multi-layer symbolic regression}: a hierarchical architecture that discovers latent intermediate symbolic functions, composing them layer by layer to handle ICC and capture hierarchical biological structure.  
  \end{enumerate}
  
  Together, their integration enables practical application to high-dimensional biomedical datasets where existing approaches fail.

  Beyond predictive performance, our goal is to demonstrate that symbolic regression can recover plausible functional relationships directly from observational data, without post hoc interpretation. We evaluate these contributions on four Feynman physics benchmarks and seven real-world datasets spanning biomedicine and education, including the Raine Study Generation 2 longitudinal cohort (Raine). On Raine, DeepPySR's multi-layer architecture recovers an interaction between polygenic risk score and early-life BMI trajectory as a latent intermediate term, pointing to a mechanistic link between genetic susceptibility and developmental growth patterns that a single-layer formulation would not expose.
  \section{Results}
  \label{sec:results}

  We evaluated DeepPySR against PySR, KAN, Random Forest, Extra Trees, XGBoost, MLP, and ElasticNet across two benchmark suites.  The first consists of four Feynman physics equations~\cite{udrescu2020ai}, which are noise-free, ground-truth-known targets for assessing exact recovery.  The second is seven real-world datasets spanning diverse and challenging conditions: regression with continuous targets (body fat, wine quality, student performance), binary classification with severe class imbalance (heart disease, stroke and diabetes), multiclass classification (diabetes), and a high-dimensional longitudinal regression with severe intraclass correlation (ICC), missingness, and genetic covariates (Raine BMI).  
  Together, these tasks stress-test DeepPySR's target failure modes: high dimensionality and ICC (DVPS), formula selection (EPS), and hierarchical structure (multi-layer). 
  Full dataset details and preprocessing protocols are described in Methods (Sec.~\ref{sec:datasets}).  Table~\ref{tab:dataset_summary} provides background information and repository locations for each real-world dataset.

  \begin{table}[h]
    \centering
    \small
    \caption{Summary of real-world benchmark datasets. $N$: number of samples; $d$: number of features (after label encoding and before SMOTE). BRFSS: CDC Behavioral Risk Factor Surveillance System. $^\dagger$Available to approved researchers via application to the Raine Study Management Committee. Full preprocessing details for all datasets are in Supplementary Note~S1.}
    \label{tab:dataset_summary}
    \resizebox{\textwidth}{!}{%
    \begin{tabular}{llrrll}
      \toprule
      Dataset & Task & $N$ & $d$ & Domain & Repository / Source \\
      \midrule
      Body Fat             & Regression              & 252          & 14  & Clinical      & UCI ML Repository \\
      Heart Disease        & Binary classification   & 303          & 13  & Clinical      & UCI ML Repository (Cleveland) \\
      Wine Quality (Red)   & Regression              & 1,599        & 11  & Food science  & UCI ML Repository \\
      Wine Quality (White) & Regression              & 4,898        & 11  & Food science  & UCI ML Repository \\
      Student Math         & Regression              & 395          & 30  & Education     & UCI ML Repository \\
      Student Portuguese   & Regression              & 649          & 30  & Education     & UCI ML Repository \\
      Stroke               & Binary classification   & 5,110        & 10  & Clinical      & UCI ML Repository \\
      Diabetes             & Multiclass (3-class)    & 253,680      & 21  & Public health & CDC BRFSS \\
      Raine BMI            & Longitudinal regression & 584--1,046$^\dagger$ & 125 & Cohort study & Raine Study, UWA \\
      \bottomrule
    \end{tabular}}
  \end{table}

  \subsection{Feynman Physics Benchmarks}

  The four benchmark equations span one to seven inputs across probability, mechanics, and electromagnetism:
  \begin{align}
    \text{I.6.2a}\;(n_f=1){:}\quad
    & f(\theta) = \frac{1}{\sqrt{2\pi}}\exp\Bigl(-\tfrac{\theta^2}{2}\Bigr)
    \label{eq:f_I6} \\
    \text{I.13.4}\;(n_f=4){:}\quad
    & f = \tfrac{1}{2}m\bigl(v_1^2+v_2^2+v_3^2\bigr)
    \label{eq:f_I13} \\
    \text{I.9.18}\;(n_f=7){:}\quad
    & f = \frac{F_1 F_2}{4\pi\varepsilon_0\bigl[(x_2\!-\!x_1)^2+(y_2\!-\!y_1)^2+(z_2\!-\!z_1)^2\bigr]}
    \label{eq:f_I9} \\
    \text{I.32.17}\;(n_f=6){:}\quad
    & f = \frac{q^2 a^2}{6\pi\varepsilon_0 c^3\bigl(1-(v/c)^2\bigr)}
    \label{eq:f_I32}
  \end{align}
  Each equation is evaluated on $N=1{,}000$ noise-free samples; benchmark metadata are provided in Supplementary Note~S1.1 (dataset descriptions and preprocessing pipelines for all datasets are in Supplementary Note~S1). Table~\ref{tab:feynman_results} reports the best R$^2$, RMSE, and formula complexity for each method; full per-model metrics and discovered formulas are in Supplementary Note~S2.

  \begin{table*}[t]
    \centering
    \caption{Feynman physics benchmark results (I.6.2a, I.13.4, I.9.18, I.32.17).  Metrics: mean $\pm$ SE across 5 folds.
    C: node count of best formula (--- for non-symbolic models).
    ``1.000'': exact recovery ($R^2 \geq 1{-}10^{-8}$, RMSE $\approx 0$).
    \textbf{Bold}: best R$^2$/RMSE per column (DeepPySR row bolded throughout).
    $^\dagger$ next to R$^2$ value: one-sided Wilcoxon signed-rank $p < 0.05$ vs.\ DeepPySR.}
    \label{tab:feynman_results}
    \resizebox{\textwidth}{!}{%
    \begin{tabular}{l rrr rrr}
      \toprule
      & \multicolumn{3}{c}{\textit{I.6.2a} ($n_f$=1): Gaussian probability}
      & \multicolumn{3}{c}{\textit{I.13.4} ($n_f$=4): Kinetic energy} \\
      \cmidrule(lr){2-4}\cmidrule(lr){5-7}
      Model & R$^2\pm$SE & RMSE$\pm$SE & C & R$^2\pm$SE & RMSE$\pm$SE & C \\
      \midrule
      \textbf{DeepPySR}      & $\mathbf{1.000}\pm0.000$ & $\mathbf{0.000}\pm0.000$ & \textbf{10}
                             & $\mathbf{1.000}\pm0.000$ & $\mathbf{0.000}\pm0.000$ & \textbf{18} \\
      PySR                   & $\mathbf{1.000}\pm0.000$ & $\mathbf{0.000}\pm0.000$ & 9
                             & $\mathbf{1.000}\pm0.000$ & $\mathbf{0.000}\pm0.000$ & 17 \\
      KAN          & $0.000^\dagger\pm0.000$ & $0.071\pm0.001$ & ---
                             & $1.000^\dagger\pm0.000$ & $0.183\pm0.013$ & --- \\
      RandomForest & $1.000^\dagger\pm0.000$ & $0.001\pm0.000$ & ---
                             & $0.903^\dagger\pm0.002$ & $8.282\pm0.207$ & --- \\
      ExtraTrees   & $0.992^\dagger\pm0.002$ & $0.006\pm0.001$ & ---
                             & $0.881^\dagger\pm0.007$ & $9.191\pm0.433$ & --- \\
      XGBoost      & $0.999^\dagger\pm0.000$ & $0.002\pm0.000$ & ---
                             & $0.980^\dagger\pm0.001$ & $3.802\pm0.099$ & --- \\
      MLP          & $0.995^\dagger\pm0.001$ & $0.005\pm0.000$ & ---
                             & $0.995^\dagger\pm0.001$ & $1.841\pm0.176$ & --- \\
      ElasticNet   & $0.390^\dagger\pm0.003$ & $0.055\pm0.001$ & ---
                             & $0.913^\dagger\pm0.005$ & $7.879\pm0.316$ & --- \\
      \midrule
      & \multicolumn{3}{c}{\textit{I.9.18} ($n_f$=7): Gravitational force}
      & \multicolumn{3}{c}{\textit{I.32.17} ($n_f$=6): Electromagnetic scattering} \\
      \cmidrule(lr){2-4}\cmidrule(lr){5-7}
      \textbf{DeepPySR}      & $0.996\pm0.000$ & $0.008\pm0.042$ & 43
                             & $1.000\pm0.000$ & $0.028\pm0.001$ & 37 \\
      PySR                   & $\mathbf{0.998}\pm0.000$ & $\mathbf{0.006}\pm0.173$ & \textbf{41}
                             & $\mathbf{1.000}\pm0.000$ & $\mathbf{0.000}\pm0.000$ & \textbf{44} \\
      KAN          & $0.000^\dagger\pm0.000$ & $0.121\pm0.003$ & ---
                             & $0.930^\dagger\pm0.040$ & $1.300\pm0.376$ & --- \\
      RandomForest & $0.675^\dagger\pm0.017$ & $0.069\pm0.002$ & ---
                             & $0.681^\dagger\pm0.025$ & $2.775\pm0.295$ & --- \\
      ExtraTrees   & $0.653^\dagger\pm0.016$ & $0.071\pm0.003$ & ---
                             & $0.657^\dagger\pm0.041$ & $2.878\pm0.353$ & --- \\
      XGBoost      & $0.914^\dagger\pm0.004$ & $0.036\pm0.001$ & ---
                             & $0.804^\dagger\pm0.031$ & $2.177\pm0.265$ & --- \\
      MLP          & $0.985^\dagger\pm0.002$ & $0.015\pm0.001$ & ---
                             & $0.944^\dagger\pm0.012$ & $1.161\pm0.211$ & --- \\
      ElasticNet   & $0.079^\dagger\pm0.008$ & $0.116\pm0.003$ & ---
                             & $0.498^\dagger\pm0.028$ & $3.485\pm0.300$ & --- \\
      \bottomrule
    \end{tabular}}
  \end{table*}

  DeepPySR and PySR both achieve exact symbolic recovery ($R^2 \geq 1-10^{-8}$, RMSE $\approx 0$) on I.6.2a (Eq.~\ref{eq:f_I6}) and I.13.4 (Eq.~\ref{eq:f_I13}) (Table~\ref{tab:feynman_results}).  For Feynman dataset I.9.18 (gravitational force; Eq.~\ref{eq:f_I9}), PySR achieves exact recovery ($R^2 = 0.998\pm0.000$, RMSE $= 0.006$, 41 nodes), while DeepPySR reaches $R^2 = 0.996\pm0.000$ with a small residual (RMSE $= 0.008$, 43 nodes). The division-by-sum-of-squared-differences structure is difficult to evolve without structural guidance, yet the approximation remains highly accurate.  For I.32.17 (Eq.~\ref{eq:f_I32}), both methods reach $R^2 = 1.000$; PySR recovers the exact formula (RMSE $\approx 0$, 44 nodes), while DeepPySR returns a structurally close approximation (RMSE $= 0.028\pm0.001$, 37 nodes), reflecting a complexity–accuracy trade-off in EPS selection.  Conventional models (XGBoost, MLP, Random Forest) achieve intermediate $R^2$ on some equations, but return no formula; ElasticNet fails on the nonlinear tasks (I.9.18: $R^2 = 0.079$).  Statistical significance is assessed with one-sided Wilcoxon signed-rank tests~\cite{wilcoxon1945individual} on the five per-fold R$^2$ values ($n=5$; minimum achievable $p=0.031$); R$^2$ values marked $^\dagger$ indicate DeepPySR significantly outperforms that model ($p<0.05$).  On I.6.2a, I.13.4, I.9.18, and I.32.17 all conventional baselines are significant ($p=0.031$); PySR is not significant on any equation, consistent with near-identical fold-level recovery.  
  DVPS/EPS ablations and Pareto-front visualisations are in Supplementary Note~S3.

  Figure~\ref{fig:convergence} shows MSE vs.\ iteration under identical settings (\texttt{adaptive\_parsimony\_scaling}$= 10$), isolating the effect of DVPS. For I.13.4 (4 variables), DeepPySR improves by three orders of magnitude, with pruning activating around iteration 50, whereas PySR stagnates.  For the Feynman dataset I.9.18 (7 variables), PySR stalls at MSE $\approx 0.013$ while DeepPySR reaches $10^{-4}$. 
  The 7-input search space benefits disproportionately from dynamic pruning. 
  On the 125-feature Raine BMI task, PySR plateaus at MSE $\approx 19$ while DeepPySR reaches $1.4$, a $13\times$ improvement, demonstrating that DVPS is essential for datasets where $d$ far exceeds the number of true causal variables.

  \begin{figure*}[!t]
    \centering
    \begin{subfigure}[b]{0.31\textwidth}
      \includegraphics[width=\linewidth]{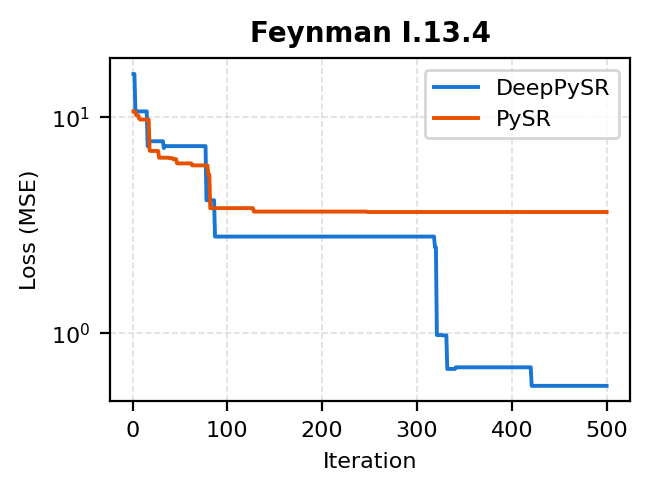}
      \caption{I.13.4 ($n_f=4$)}
    \end{subfigure}\hfill
    \begin{subfigure}[b]{0.31\textwidth}
      \includegraphics[width=\linewidth]{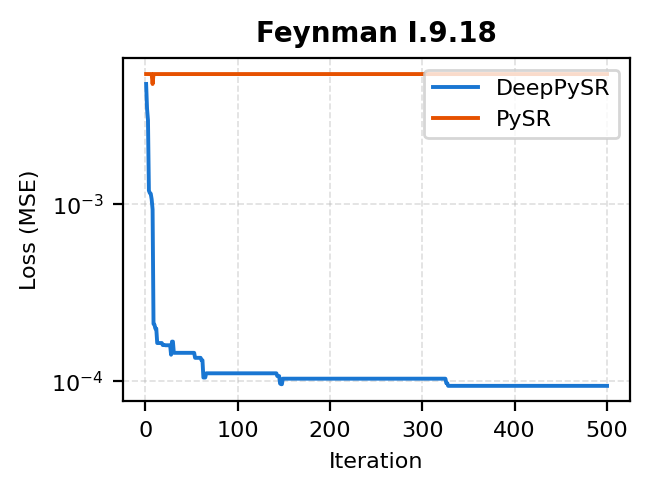}
      \caption{I.9.18 gravitational force ($n_f=7$)}
    \end{subfigure}\hfill
    \begin{subfigure}[b]{0.31\textwidth}
      \includegraphics[width=\linewidth]{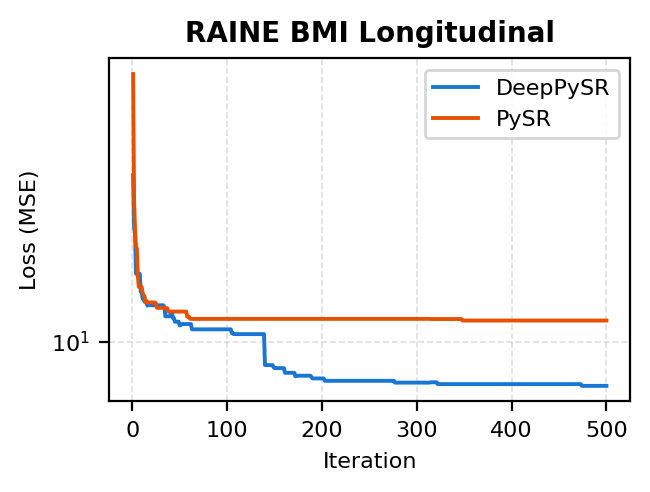}
      \caption{Raine BMI ($n_f=125$)}
    \end{subfigure}

    \vspace{1em}

    \begin{subfigure}[b]{0.46\textwidth}
      \includegraphics[width=\linewidth]{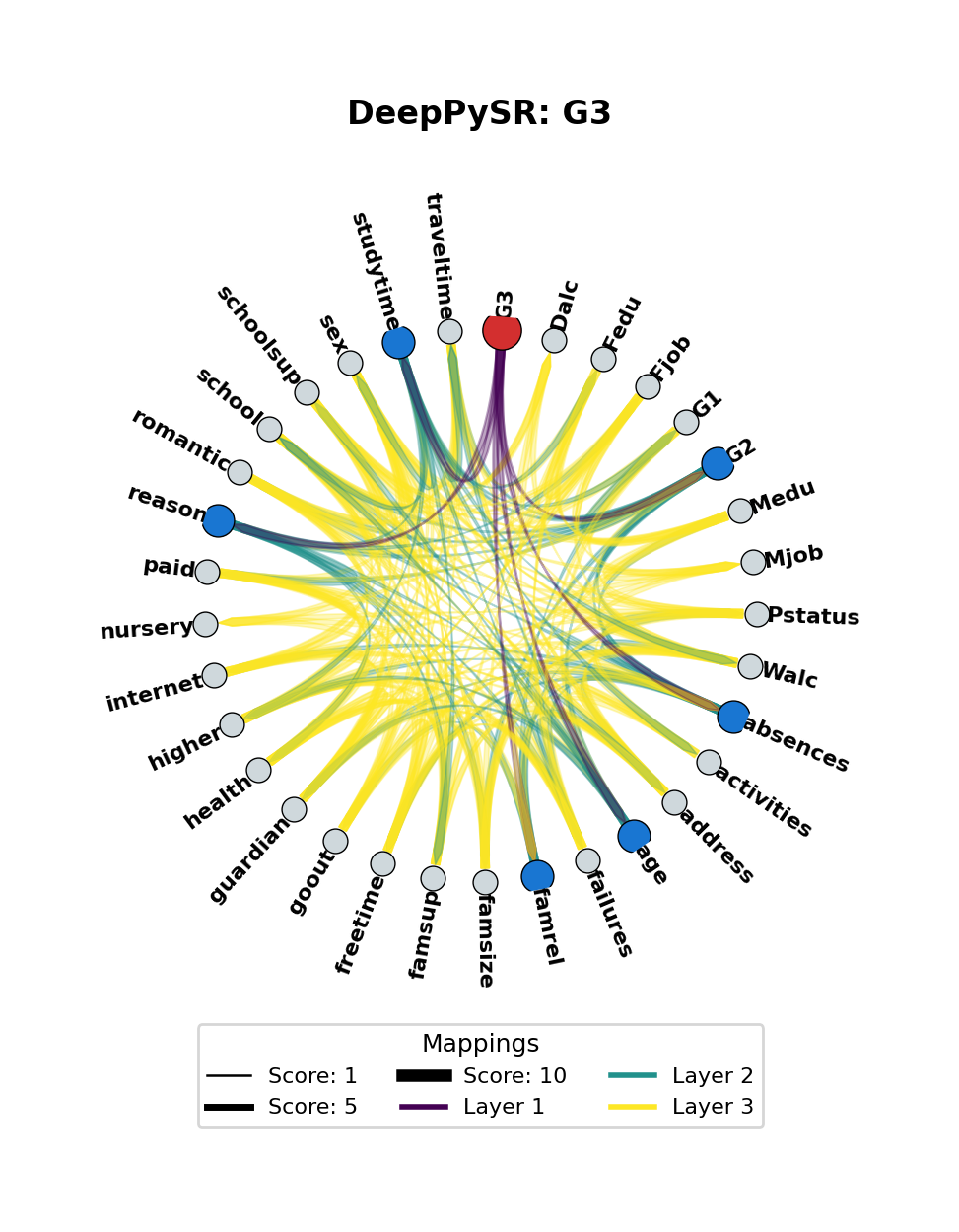}
      \caption{Student performance (Math)}
    \end{subfigure}\hfill
    \begin{subfigure}[b]{0.5\textwidth}
      \includegraphics[width=\linewidth]{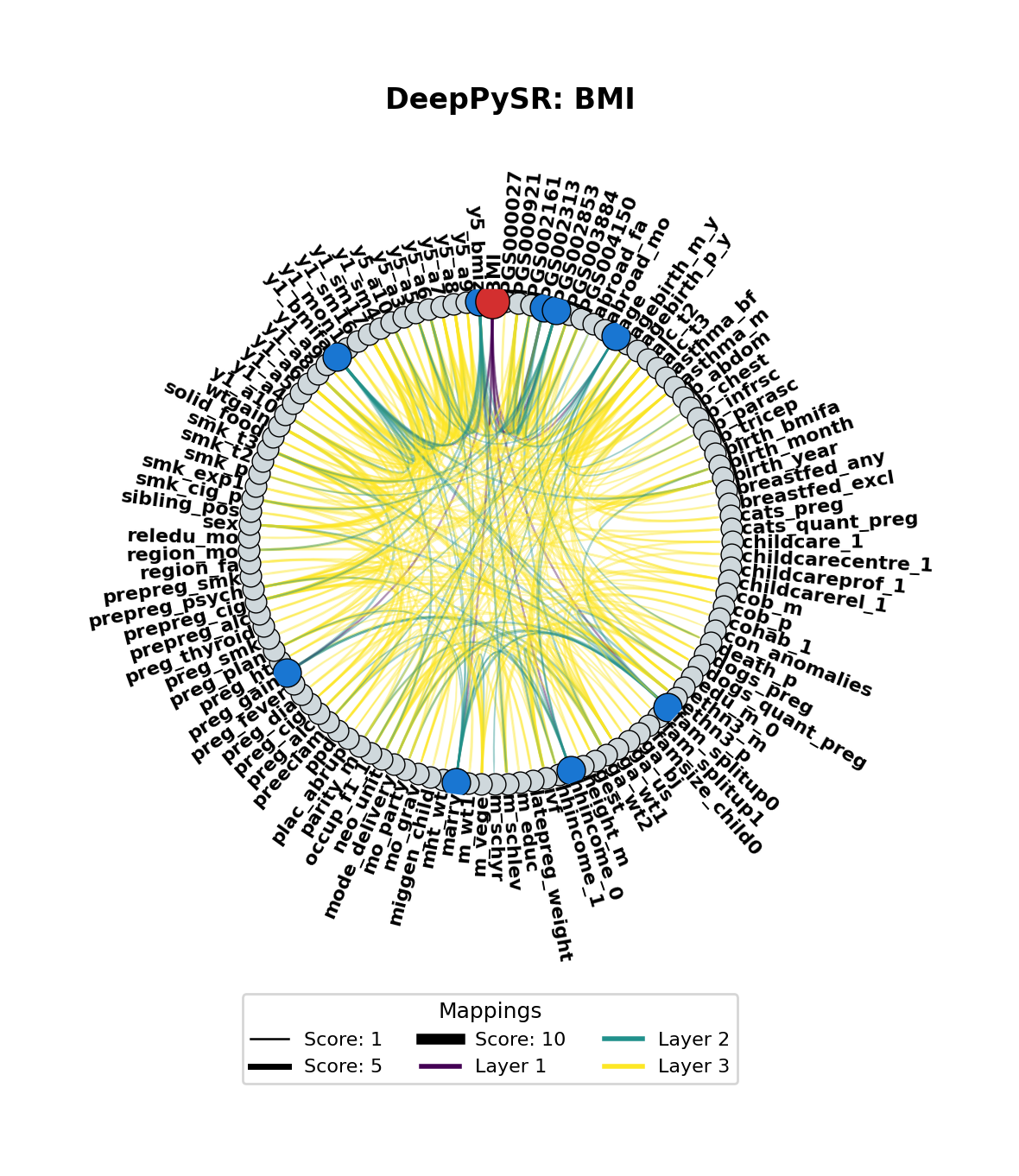}
      \caption{Raine BMI longitudinal}
    \end{subfigure}

    \caption{\textbf{(a--c) Convergence curves} (MSE vs.\ iteration, log scale).
    Both models use identical hyperparameters (\texttt{adaptive\_parsimony\_scaling}$=10$,
      all others at default); losses differ from the grid-searched results in
      Table~\ref{tab:feynman_results}.
      \textbf{(d--e) Relationship circle plots.}
      Arc colour encodes SR layer (Layer~1: purple; Layer~2: cyan; Layer~3: yellow);
      arc thickness encodes connection score.
      See main text for quantitative analysis.}
    \label{fig:convergence}
  \end{figure*}

  \subsection{Regression Datasets}

  Table~\ref{tab:regression_results} compares all models on body fat (252 samples, 14 anthropometric features; clean, high ICC), wine quality (red: 1,599 samples; white: 4,898 samples; 11 features; noisy subjective ratings with correlated chemical predictors), and student performance (Math: 395 samples; Portuguese: 649 samples; 30 features; strong prior-grade ICC, social covariates).  These datasets span from balanced, well-conditioned regression to high-ICC scenarios where correlated predictors confound variable selection.

  \begin{table*}[t]
    \centering
    \caption{Regression results.  Metrics: mean $\pm$ SE across 5 folds.
    \textbf{Bold}: best per column (DeepPySR row bolded throughout).
    \underline{Underline}: DeepPySR interpretable formula row.
    $^\dagger$ next to R$^2$ value: one-sided Wilcoxon signed-rank $p < 0.05$ vs.\ DeepPySR.
    MAE in Supplementary Note~S2.}
    \label{tab:regression_results}
    \resizebox{\textwidth}{!}{%
    \begin{tabular}{l rr rr rr}
      \toprule
      & \multicolumn{2}{c}{\textit{Body Fat} (252, 14f)}
      & \multicolumn{2}{c}{\textit{Wine Red} (1{,}599, 11f)}
      & \multicolumn{2}{c}{\textit{Wine White} (4{,}898, 11f)} \\
      \cmidrule(lr){2-3}\cmidrule(lr){4-5}\cmidrule(lr){6-7}
      Model & R$^2\pm$SE & RMSE$\pm$SE & R$^2\pm$SE & RMSE$\pm$SE
            & R$^2\pm$SE & RMSE$\pm$SE \\
      \midrule
      \textbf{DeepPySR (best)}

        & $\mathbf{0.794\pm0.028}$ & $\mathbf{3.793\pm0.084}$
        & $\mathbf{0.412\pm0.031}$ & $\mathbf{0.619\pm0.014}$
        & $0.349\pm0.009$ & $0.715\pm0.005$ \\
      \underline{DeepPySR (interp.)}
        & $\underline{0.772\pm0.028}$ & $\underline{3.986\pm0.084}$
        & $\underline{0.363\pm0.031}$ & $\underline{0.644\pm0.014}$
        & $\underline{0.301\pm0.009}$ & $\underline{0.740\pm0.005}$ \\
      PySR
        & $0.702\pm0.014$ & $4.558\pm0.128$
        & $0.308^\dagger\pm0.048$ & $0.671\pm0.021$
        & $0.186^\dagger\pm0.006$ & $0.799\pm0.003$ \\
      XGBoost
        & $0.650\pm0.031$ & $4.939\pm0.200$
        & $0.394\pm0.025$ & $0.628\pm0.013$
        & $0.412\pm0.008$ & $0.679\pm0.005$ \\
      RandomForest
        & $0.671\pm0.031$ & $4.788\pm0.179$
        & $0.405\pm0.011$ & $0.623\pm0.005$
        & $0.351\pm0.009$ & $0.713\pm0.005$ \\
      ExtraTrees
        & $0.646^\dagger\pm0.032$ & $4.971\pm0.126$
        & $0.368\pm0.009$ & $0.642\pm0.004$
        & $0.302\pm0.006$ & $0.740\pm0.003$ \\
      MLP
        & $0.651\pm0.034$ & $4.937\pm0.176$
        & $0.372\pm0.014$ & $0.640\pm0.006$
        & $\mathbf{0.457\pm0.011}$ & $\mathbf{0.652\pm0.007}$ \\
      ElasticNet
        & $0.693\pm0.022$ & $4.629\pm0.152$
        & $0.328\pm0.008$ & $0.662\pm0.003$
        & $0.237^\dagger\pm0.005$ & $0.773\pm0.003$ \\
      KAN
        & $0.128^\dagger\pm0.093$ & $7.799\pm0.550$
        & $0.346\pm0.013$ & $0.653\pm0.006$
        & $0.343\pm0.010$ & $0.718\pm0.005$ \\
      \midrule
      & \multicolumn{2}{c}{\textit{Student Mat} (395, 30f)}
      & \multicolumn{2}{c}{\textit{Student Por} (649, 30f)} \\
      \cmidrule(lr){2-3}\cmidrule(lr){4-5}
      \textbf{DeepPySR (best)}
        & $\mathbf{0.964\pm0.004}$ & $\mathbf{0.866\pm0.049}$
        & $\mathbf{0.924\pm0.010}$ & $\mathbf{0.888\pm0.054}$ & & \\
      \underline{DeepPySR (interp.)}
        & $\underline{0.943\pm0.004}$ & $\underline{1.088\pm0.049}$
        & $\underline{0.892\pm0.010}$ & $\underline{1.059\pm0.054}$ & & \\
      PySR      & $0.948\pm0.019$ & $1.043\pm0.159$ & $0.892^\dagger\pm0.008$ & $1.063\pm0.057$ & & \\
      XGBoost   & $0.833^\dagger\pm0.013$ & $1.870\pm0.066$ & $0.812^\dagger\pm0.015$ & $1.400\pm0.057$ & & \\
      RandomForest & $0.865^\dagger\pm0.014$ & $1.680\pm0.079$ & $0.836^\dagger\pm0.019$ & $1.308\pm0.076$ & & \\
      ExtraTrees & $0.828^\dagger\pm0.008$ & $1.897\pm0.072$ & $0.815^\dagger\pm0.010$ & $1.389\pm0.057$ & & \\
      MLP        & $0.764^\dagger\pm0.030$ & $2.224\pm0.149$ & $0.798^\dagger\pm0.017$ & $1.450\pm0.074$ & & \\
      ElasticNet & $0.815^\dagger\pm0.006$ & $1.966\pm0.060$ & $0.841^\dagger\pm0.017$ & $1.289\pm0.088$ & & \\
      KAN        & $0.584^\dagger\pm0.038$ & $2.951\pm0.141$ & $0.420^\dagger\pm0.095$ & $2.459\pm0.172$ & & \\
      \bottomrule
    \end{tabular}}
  \end{table*}

  \textbf{Body fat.}  After deliberately removing the density (0.99 correlated to bodyfat), DeepPySR achieves $R^2 = 0.794 \pm 0.028$, improving over PySR by $+0.092$ and over the best conventional model, ElasticNet ($R^2 = 0.693 \pm 0.022$), by $+0.101$; here and throughout Results, $+x$ ($-x$) denotes that DeepPySR's metric is $x$ higher (lower) than the comparator.  Without pruning, PySR's evolutionary search converges on the high-weight Weight--Abdomen collinear cluster, saturating complexity without generalising.  

  DVPS progressively penalises redundancy, retaining broader anthropometric diversity.  
  The interpretable formula ($R^2 = 0.772$, 34 nodes) retains 97\% of the best configuration's $R^2$:
  \begin{equation}
    \hat{y} = \mathrm{Abdo} - \frac{543.39\,\bigl(-\mathrm{Wrist}+\sin(11.63\,W)-\sin(738.74\,\mathrm{Neck}+\sin(0.39\,W))\bigr)}{W - 515.23} - 43.49,
    \label{eq:bodyfat}
  \end{equation}
  where $W$=Weight, Abdo=Abdomen circumference, Neck=Neck circumference, Wrist=Wrist circumference.  The formula is centred on the abdominal circumference, the primary variable in established clinical body fat equations, with wrist and neck circumferences as secondary terms, consistent with published regression models for this dataset~\cite{johnson1996fitting}.  Wilcoxon tests confirm DeepPySR significantly outperforms ExtraTrees and KAN ($p=0.031$).  PySR is not significantly worse ($p=0.219$): the $+0.092$ R$^2$ improvement is real, but the 5-fold variance is too large for the test to reliably detect it at $n=5$.

  \textbf{Wine quality.}  DeepPySR achieves $R^2 = 0.412$ (red), outperforming PySR ($+0.104$) and Random Forest ($+0.007$, best conventional model at $R^2 = 0.405$). 
  DVPS isolates volatile acidity, alcohol and sulphates, excluding correlated confounders (density, pH, total sulphur dioxide) that add complexity without improving generalisation. 
  The red wine interpretable formula ($R^2 = 0.363$, 23 nodes; VA=volatile acidity, Sulph=sulphates):
  \begin{equation}
    \hat{y} = -\mathrm{VA} + 1.91\sqrt{\mathrm{Alcohol} - \sin\!\left(\tfrac{\mathrm{pH}}{\mathrm{Sulph}}-5.63\right)}.
    \label{eq:wine_red}
  \end{equation}
  White wine is inherently harder, ratings reflect subjective sensory evaluation, and the target distribution is noisy; MLP achieves the best overall $R^2 = 0.457$, while DeepPySR ($R^2 = 0.349$) produces the only interpretable formula (23 nodes; FA=fixed acidity, RS=residual sugar, FSO$_2$=free sulphur dioxide, Chl=chlorides):
  \begin{equation}
    \hat{y} = -\frac{1.17\,\mathrm{FA}}{\mathrm{RS}+\mathrm{FSO}_2} + \mathrm{VA}\cdot(\mathrm{Alcohol}-12.4) - \mathrm{Chl} + 6.7,
    \label{eq:wine_white}
  \end{equation}
  retaining the acidity–alcohol–sulphate triad.  
  The white wine gap is consistent across symbolic methods, reflecting interpretable models' limits when signal-to-noise is low. 
  Wilcoxon signed-rank tests confirm DeepPySR significantly outperforms PySR on red wine ($p=0.031$); on white wine, DeepPySR significantly outperforms ElasticNet and PySR ($p=0.031$).  No conventional model is significantly worse than DeepPySR on either wine type, as tree-based methods and MLP exploit correlated features more effectively at this sample size.

  \textbf{Student performance.}  DeepPySR achieves $R^2 = 0.964 \pm 0.004$ (Math) and $0.924 \pm 0.010$ (Portuguese), substantially outperforming all baselines, including PySR ($R^2 = 0.948 \pm 0.019$ and $0.892 \pm 0.008$), XGBoost ($0.833 \pm 0.013$ and $0.812 \pm 0.015$), and MLP ($0.764 \pm 0.030$ and $0.798 \pm 0.017$).  
  Strong ICC among G1, G2, and G3 causes standard methods to collapse onto a single grade predictor or to overfit social covariates. 
  The multi-layer architecture selects G2 as the Layer-1 anchor and uses G1 plus social variables in outer layers. 
  In the circle plot (Fig.~\ref{fig:convergence}d), G2 is selected via a thick Layer-1 arc as the dominant predictor of G3, with social variables appearing in Layer 2.  
  The formula captures that prior academic trajectory is the strongest predictor of final grade, with social risk factors adding independent signal. 
  
\bigskip
\noindent\textbf{Math}
\quad
($R^2=0.943$, 23 nodes; $G_2$=prior term grade, abs=absences,
st=studytime, rsn=reason, trvl=traveltime, Pst=Pstatus, highr=higher)
\begin{equation}
  \hat{y} =
  \begin{cases}
    G_2 & \text{if } \mathrm{abs} + \sin(0.41\,\mathrm{st} - 0.41\,G_2 + \sin(\mathrm{rsn}\cdot(\mathrm{trvl}+\mathrm{highr}+\mathrm{Pst}))) + 0.66 > 0 \\[6pt]
    0 & \text{otherwise}
  \end{cases}
  \label{eq:student_mat}
\end{equation}

\noindent\textbf{Portuguese}
\quad
($R^2=0.892$, 24 nodes; sch=school, act=activities)
\begin{equation}
  \hat{y} =
  \begin{cases}
    0.96\,(G_2 + 0.93) & \text{if } \mathrm{sch} + \mathrm{abs} - \sin\!\left(\sqrt{G_2 + e^{\,\mathrm{rsn}-\mathrm{act}}}\right) > 0 \\[6pt]
    0 & \text{otherwise}
  \end{cases}
  \label{eq:student_por}
\end{equation}

  Wilcoxon signed-rank tests confirm DeepPySR significantly outperforms all conventional baselines on both subjects ($p=0.031$).  On Math, PySR is not significant ($p=0.375$): both SR methods achieve high R$^2$, and the fold margins do not consistently favour one over the other.  On Portuguese, DeepPySR significantly outperforms even PySR ($p=0.031$), confirming that DVPS-aided search generalises more reliably across all five folds on this cohort.

  \subsection{Classification Datasets}

  Table~\ref{tab:classification_results} compares models on three classification tasks representing distinct challenge types: heart disease (303 samples, 13 clinical features; balanced binary classification --- 54\% positive), stroke (43,400 samples, 11 features; severe class imbalance --- 4.9\% positive --- binary), and diabetes (253,680 samples, 21 features; three-class: no diabetes\,/\,prediabetes\,/\,diabetes; class distribution 84.2\%\,/\,1.8\%\,/\,13.9\% --- heavy underrepresentation of the prediabetes class).  
  SR formula outputs are rounded to the nearest integer and clipped to valid labels (\{0, 1\} for binary tasks; \{0, 1, 2\} for diabetes).
  Baselines for stroke and diabetes use SMOTE oversampling; DeepPySR and PySR do not, to test whether the formula structure alone can handle imbalance.

  \begin{table*}[t]
    \centering
    \caption{Classification results.  Metrics: mean $\pm$ SE across 5 folds (F1 listed first).
    Datasets are grouped by row block.
    Stroke and Diabetes have severe class imbalance;
    Diabetes is multiclass (no diabetes/prediabetes/diabetes).
    SMOTE applied to baselines only; DeepPySR and PySR evaluated without SMOTE.
    \textbf{Bold}: best per column (DeepPySR row bolded throughout).
    \underline{Underline}: DeepPySR interpretable formula row.
    $^\dagger$ next to F1 value: one-sided Wilcoxon signed-rank $p < 0.05$ vs.\ DeepPySR.}
    \label{tab:classification_results}
    \begin{tabular}{l rrrr}
      \toprule
      Model & F1$\pm$SE & Acc.$\pm$SE & Prec.$\pm$SE & Rec.$\pm$SE \\
      \midrule
      \multicolumn{5}{l}{\textit{Heart Disease} (303 samples, 13 features, binary, 54\% positive)} \\
      \midrule
      \textbf{DeepPySR (best)}       & $\mathbf{0.898\pm0.016}$ & $\mathbf{0.909\pm0.012}$ & $\mathbf{0.930\pm0.011}$ & $\mathbf{0.869\pm0.036}$ \\
      \underline{DeepPySR (interp.)} & $\underline{0.879\pm0.016}$ & $\underline{0.892\pm0.012}$ & $\underline{0.913\pm0.011}$ & $\underline{0.847\pm0.036}$ \\
      PySR               & $0.787^\dagger\pm0.029$ & $0.805\pm0.030$ & $0.793\pm0.047$ & $0.781\pm0.019$ \\
      ExtraTrees         & $0.805^\dagger\pm0.021$ & $0.828\pm0.019$ & $0.847\pm0.034$ & $0.766\pm0.026$ \\
      RandomForest       & $0.789^\dagger\pm0.026$ & $0.815\pm0.021$ & $0.831\pm0.025$ & $0.752\pm0.038$ \\
      XGBoost            & $0.775^\dagger\pm0.022$ & $0.795\pm0.019$ & $0.784\pm0.024$ & $0.766\pm0.032$ \\
      MLP                & $0.774^\dagger\pm0.024$ & $0.791\pm0.020$ & $0.774\pm0.018$ & $0.774\pm0.035$ \\
      KAN                & $0.810^\dagger\pm0.029$ & $0.818\pm0.027$ & $0.782\pm0.033$ & $0.839\pm0.037$ \\
      \midrule
      \multicolumn{5}{l}{\textit{Stroke} (43{,}400 samples, 11 features, binary, 4.9\% positive; SMOTE on baselines)} \\
      \midrule
      \textbf{DeepPySR (best)}       & $\mathbf{0.257\pm0.005}$ & $0.865\pm0.005$ & $\mathbf{0.168\pm0.004}$ & $0.550\pm0.024$ \\
      \underline{DeepPySR (interp.)} & $\underline{0.257\pm0.005}$ & $\underline{0.865\pm0.005}$ & $\underline{0.168\pm0.004}$ & $\underline{0.550\pm0.024}$ \\
      PySR                         & $0.082\pm0.016$ & $\mathbf{0.936\pm0.005}$ & $0.106\pm0.010$ & $0.067\pm0.038$ \\
      ExtraTrees         & $0.202^\dagger\pm0.006$ & $0.741\pm0.011$ & $0.116\pm0.004$ & $\mathbf{0.770\pm0.023}$ \\
      RandomForest       & $0.187^\dagger\pm0.015$ & $0.756\pm0.013$ & $0.109\pm0.009$ & $0.656\pm0.059$ \\
      XGBoost            & $0.145^\dagger\pm0.009$ & $0.919\pm0.003$ & $0.131\pm0.009$ & $0.163\pm0.011$ \\
      MLP                & $0.104^\dagger\pm0.013$ & $0.884\pm0.003$ & $0.078\pm0.009$ & $0.158\pm0.022$ \\
      KAN                & $0.192^\dagger\pm0.004$ & $0.728\pm0.022$ & $0.110\pm0.004$ & $0.756\pm0.048$ \\
      \midrule
      \multicolumn{5}{l}{\textit{Diabetes} (253{,}680 samples, 21 features, 3-class; SMOTE on baselines)} \\
      \midrule
      \textbf{DeepPySR (best)}       & $0.350\pm0.001$ & $0.646\pm0.001$ & $0.527\pm0.002$ & $\mathbf{0.466\pm0.002}$ \\
      \underline{DeepPySR (interp.)} & $\underline{0.343\pm0.001}$ & $\underline{0.649\pm0.001}$ & $\underline{0.531\pm0.002}$ & $\underline{0.463\pm0.002}$ \\
      PySR               & $0.311^\dagger\pm0.000$ & $0.738\pm0.000$ & $0.305\pm0.000$ & $0.389\pm0.002$ \\
      ExtraTrees         & $0.305^\dagger\pm0.000$ & $0.842\pm0.000$ & $\mathbf{0.614\pm0.067}$ & $0.333\pm0.000$ \\
      RandomForest       & $0.319^\dagger\pm0.001$ & $0.844\pm0.000$ & $0.505\pm0.003$ & $0.340\pm0.001$ \\
      XGBoost            & $\mathbf{0.401\pm0.001}$ & $\mathbf{0.850\pm0.000}$ & $0.475\pm0.002$ & $0.389\pm0.001$ \\
      MLP                & $0.395\pm0.002$ & $0.847\pm0.000$ & $0.500\pm0.023$ & $0.385\pm0.001$ \\
      KAN                & $0.305^\dagger\pm0.000$ & $0.842\pm0.000$ & $0.281\pm0.000$ & $0.333\pm0.000$ \\
      \bottomrule
    \end{tabular}
  \end{table*}

  \textbf{Heart disease.}  DeepPySR achieves F1 $= 0.898$, outperforming PySR ($+0.111$) and Extra Trees (F1 $= 0.805$, best conventional model, $+0.093$).  A precision of $0.930$ is the highest across all models, which is clinically critical in a setting where false positives trigger unnecessary invasive investigation.  The discovered formula identifies a functional interaction between chest pain type and the ST-slope, refined by the weighting of secondary risk factors (thalassemia type [thal], sex, cholesterol, number of major vessels coloured by fluoroscopy [ca], exercise-induced angina [exang], fasting blood sugar, age).  The interpretable formula (F1 $= 0.879$, 34 nodes; cp=chest pain type, slope=slope of peak exercise ST segment, ca=number of major vessels colored by fluoroscopy, op=oldpeak ST depression, chol=cholesterol, thal=thalassemia type) retains 97.9\% of the best configuration's F1, confirming that EPS selects a formula close to the Pareto-optimal accuracy–complexity frontier:
  \begin{equation}
    \hat{y} =
      \begin{cases}
        \sin\Bigl(2.06\sin\bigl((\sin(3.18\,\mathrm{slope}\cdot\mathrm{cp})+0.19)\\
        \quad\cdot(\mathrm{sex}\cdot\mathrm{ca}+0.23\,\mathrm{op}-0.23\sin(0.98\,\mathrm{chol})+\sin(\mathrm{thal}+0.25))\bigr)\Bigr)
        & \text{if } \mathrm{op} > 0 \\[6pt]
        \sin\Bigl(2.06\sin\bigl((\sin(3.18\,\mathrm{slope}\cdot\mathrm{cp})+0.19)\\
        \quad\cdot(\mathrm{sex}\cdot\mathrm{ca}+0.23\,\mathrm{op}-0.23\sin(0.98\,\mathrm{chol})+\sin(0.25))\bigr)\Bigr)
        & \text{otherwise}
      \end{cases}
    \label{eq:heart}
  \end{equation}
  All variables selected by DeepPySR are established coronary artery disease risk factors~\cite{detrano1989international}, providing direct clinical interpretability.  Wilcoxon signed-rank tests confirm DeepPySR significantly outperforms all baselines, including PySR, on F1 ($p=0.031$), reflecting a consistent per-fold advantage on this balanced binary task.

  \textbf{Stroke.}  Severe class imbalance (4.9\% positive) causes most models to degenerate: PySR reaches 93.6\% accuracy by near-constant prediction of the majority class (F1 $= 0.082$, recall $= 0.067$).  Tree ensembles with SMOTE recover recall (Extra Trees: recall $= 0.770$) but at the cost of accuracy ($74\%$), producing high false-positive rates unsuitable for clinical screening. DeepPySR achieves recall $= 0.550$ and F1 $= 0.257$ at 86.5\% accuracy \emph{without} SMOTE, identifying over half of all stroke cases while avoiding the accuracy collapse of SMOTE-based methods.  The interpretable formula (F1 $= 0.257$, 27 nodes; wt=work\_type, hyp=hypertension, res=Residence\_type) captures work type as a piecewise gate: the formula only assigns positive risk when the work type falls above a threshold, then applies a sinusoidal combination of hypertension, residence type, and age, matching the known interaction between occupational, vascular, and demographic stroke risk factors:
  \begin{equation}
    \hat{y} =
      \begin{cases}
        \sin\Bigl((\sin(3.14\,\mathrm{hyp})+\sin(3.14\,\mathrm{res}))^{0.25}+\sin(0.02\,\mathrm{age}-\sin(31.44\,\mathrm{age}))\Bigr)
        & \text{if } \mathrm{wt} > 0 \\[6pt]
        0 & \text{otherwise}
      \end{cases}
    \label{eq:stroke}
  \end{equation}
  The advantage over PySR (F1: +0.175) reflects DeepPySR’s ability to handle imbalance through formula structure rather than resampling.  
  Wilcoxon signed-rank tests confirm DeepPySR significantly outperforms all conventional baselines ($p=0.031$).  Against PySR the test does not reach significance ($p=0.0625$): PySR's degenerate near-constant predictions vary by fold in a way that produces one fold where the signed-rank sum falls below the critical value, yet the practical F1 gap of $+0.175$ is substantial and clinically meaningful.

  \textbf{Diabetes.}  XGBoost achieves the best overall F1 $= 0.401$ (with SMOTE).  DeepPySR's interpretable formula (F1 $= 0.343$, 24 nodes) outperforms PySR ($0.311$) and combines established risk factors, such as general health status (GH), high blood pressure (HBP), high cholesterol (HC), heavy alcohol consumption (HA), age, and BMI, in a compact product form:
  \begin{equation}
    \hat{y} = \mathrm{GH}\cdot(0.12\,\mathrm{HBP}+0.12\,\mathrm{HC}-0.12\,\mathrm{HA}+0.02\,\mathrm{Age})\cdot\sin(0.04\,\mathrm{BMI}-0.48).
    \label{eq:diabetes}
  \end{equation}
  The product structure is notable: it implies that risk is multiplicative across comorbidities rather than additive, consistent with epidemiological models of metabolic syndrome.  The prediabetes class remains the hardest to identify for all methods, reflecting the fundamental challenge of a minority class embedded between two larger ones.  Wilcoxon signed-rank tests confirm DeepPySR significantly outperforms ExtraTrees, RandomForest, KAN, and PySR ($p=0.031$); MLP and XGBoost are not significantly worse, as SMOTE resampling enables them to win some folds on the prediabetes class where DeepPySR (without SMOTE) does not dominate every fold.

  \subsection{Raine BMI Longitudinal}

  Table~\ref{tab:bmi_results} reports results for the Raine Study Generation 2 longitudinal BMI prediction task~\cite{straker2017raine}, which is the most demanding benchmark in this study.  With 125 candidate features (anthropometric measures, polygenic risk scores, sociodemographic and perinatal variables), severe longitudinal missingness (10--50\% attrition per follow-up wave), seven target ages, and high ICC across anthropometric and genetic predictors, this dataset represents a realistic clinical prediction problem that overwhelms standard SR and remains highly challenging for conventional baselines at later follow-up ages.  SE is constant across ages for each model because it derives from the same 5-fold longitudinal evaluation.

  \begin{table*}[h]
    \centering
    \caption{Raine BMI longitudinal prediction at ages 8--27.
    R$^2$ and RMSE: mean $\pm$ SE across 5 folds.
    SE is constant across ages for each model (single longitudinal evaluation).
    \textbf{Bold}: best R$^2$ per age.  \underline{Underline}: best interpretable model.
    $^\dagger$ next to R$^2$ value: one-sided Wilcoxon signed-rank $p < 0.05$ vs.\ DeepPySR at that age.}
    \label{tab:bmi_results}
    \resizebox{\textwidth}{!}{%
    \begin{tabular}{l rr rr rr}
      \toprule
      & \multicolumn{2}{c}{\textit{Age 8}}
      & \multicolumn{2}{c}{\textit{Age 10}}
      & \multicolumn{2}{c}{\textit{Age 14}} \\
      \cmidrule(lr){2-3}\cmidrule(lr){4-5}\cmidrule(lr){6-7}
      Model & R$^2\pm$SE & RMSE$\pm$SE & R$^2\pm$SE & RMSE$\pm$SE & R$^2$$\pm$SE & RMSE$\pm$SE \\
      \midrule
      \textbf{DeepPySR (best)}
        & $0.727\pm0.012$ & $1.242\pm0.071$
        & $\mathbf{0.604}\pm0.012$ & $\mathbf{2.083}\pm0.071$
        & $\mathbf{0.533}\pm0.012$ & $\mathbf{2.721}\pm0.071$ \\
      \underline{DeepPySR (interp.)}
        & $\underline{0.674}\pm0.012$ & $\underline{1.358}\pm0.071$
        & $\underline{0.624}\pm0.012$ & $\underline{2.029}\pm0.071$
        & $\underline{0.560}\pm0.012$ & $\underline{2.642}\pm0.071$ \\
      PySR
        & $0.665\pm0.009$ & $1.378\pm0.069$
        & $0.499\pm0.009$ & $2.344\pm0.069$
        & $0.395\pm0.009$ & $3.098\pm0.069$ \\
      RandomForest
        & $\mathbf{0.765}\pm0.025$ & $\mathbf{1.153}\pm0.234$
        & $0.571^\dagger\pm0.025$ & $2.168\pm0.234$
        & $0.398\pm0.025$ & $3.089\pm0.234$ \\
      ExtraTrees
        & $0.752^\dagger\pm0.021$ & $1.184\pm0.219$
        & $0.536^\dagger\pm0.021$ & $2.255\pm0.219$
        & $0.405\pm0.021$ & $3.071\pm0.219$ \\
      XGBoost
        & $0.603^\dagger\pm0.022$ & $1.499\pm0.223$
        & $0.493^\dagger\pm0.022$ & $2.358\pm0.223$
        & $0.402^\dagger\pm0.022$ & $3.081\pm0.223$ \\
      ElasticNet
        & $0.330^\dagger\pm0.009$ & $1.947\pm0.167$
        & $0.564\pm0.009$ & $2.187\pm0.167$
        & $0.436\pm0.009$ & $2.990\pm0.167$ \\
      MLP
        & $0.455^\dagger\pm0.009$ & $1.755\pm0.144$
        & $0.377^\dagger\pm0.009$ & $2.613\pm0.144$
        & $0.336^\dagger\pm0.009$ & $3.247\pm0.144$ \\
      KAN
        & $0.000^\dagger\pm0.000$ & $6.358\pm0.884$
        & $0.000^\dagger\pm0.000$ & $7.237\pm0.884$
        & $0.000^\dagger\pm0.000$ & $8.656\pm0.884$ \\
      \midrule
      & \multicolumn{2}{c}{\textit{Age 17}}
      & \multicolumn{2}{c}{\textit{Age 20}}
      & \multicolumn{2}{c}{\textit{Age 23}} \\
      \cmidrule(lr){2-3}\cmidrule(lr){4-5}\cmidrule(lr){6-7}
      \textbf{DeepPySR (best)}
        & $\mathbf{0.460}\pm0.012$ & $\mathbf{2.945}\pm0.071$
        & $\mathbf{0.462}\pm0.012$ & $\mathbf{3.585}\pm0.071$
        & $\mathbf{0.461}\pm0.012$ & $\mathbf{3.661}\pm0.071$ \\
      \underline{DeepPySR (interp.)}
        & $\underline{0.429}\pm0.012$ & $\underline{3.029}\pm0.071$
        & $\underline{0.428}\pm0.012$ & $\underline{3.697}\pm0.071$
        & $\underline{0.408}\pm0.012$ & $\underline{3.837}\pm0.071$ \\
      PySR
        & $0.325\pm0.009$ & $3.293\pm0.069$
        & $0.265\pm0.009$ & $4.191\pm0.069$
        & $0.232\pm0.009$ & $4.369\pm0.069$ \\
      RandomForest
        & $0.328^\dagger\pm0.025$ & $3.285\pm0.234$
        & $0.295\pm0.025$ & $4.105\pm0.234$
        & $0.243\pm0.025$ & $4.339\pm0.234$ \\
      ExtraTrees
        & $0.346\pm0.021$ & $3.241\pm0.219$
        & $0.309\pm0.021$ & $4.065\pm0.219$
        & $0.275\pm0.021$ & $4.246\pm0.219$ \\
      XGBoost
        & $0.291^\dagger\pm0.022$ & $3.374\pm0.223$
        & $0.309\pm0.022$ & $4.065\pm0.223$
        & $0.202^\dagger\pm0.022$ & $4.456\pm0.223$ \\
      ElasticNet
        & $0.362\pm0.009$ & $3.202\pm0.167$
        & $0.309^\dagger\pm0.009$ & $4.064\pm0.167$
        & $0.299^\dagger\pm0.009$ & $4.176\pm0.167$ \\
      MLP
        & $0.270^\dagger\pm0.009$ & $3.424\pm0.144$
        & $0.226^\dagger\pm0.009$ & $4.303\pm0.144$
        & $0.121^\dagger\pm0.009$ & $4.675\pm0.144$ \\
      KAN
        & $0.000^\dagger\pm0.000$ & $8.857\pm0.884$
        & $0.000^\dagger\pm0.000$ & $9.798\pm0.884$
        & $0.000^\dagger\pm0.000$ & $10.135\pm0.884$ \\
      \midrule
      & \multicolumn{2}{c}{\textit{Age 27}} & & & & \\
      \cmidrule(lr){2-3}
      \textbf{DeepPySR (best)}       & $\mathbf{0.425}\pm0.012$ & $\mathbf{4.189}\pm0.071$ & & & & \\
      \underline{DeepPySR (interp.)} & $\underline{0.354}\pm0.012$ & $\underline{4.440}\pm0.071$ & & & & \\
      PySR                 & $0.210\pm0.009$ & $4.908\pm0.069$ & & & & \\
      RandomForest         & $0.232^\dagger\pm0.025$ & $4.841\pm0.234$ & & & & \\
      ExtraTrees           & $0.236\pm0.021$ & $4.828\pm0.219$ & & & & \\
      XGBoost              & $0.175^\dagger\pm0.022$ & $5.018\pm0.223$ & & & & \\
      ElasticNet           & $0.266^\dagger\pm0.009$ & $4.732\pm0.167$ & & & & \\
      MLP                  & $0.118^\dagger\pm0.009$ & $5.187\pm0.144$ & & & & \\
      KAN                  & $0.000^\dagger\pm0.000$ & $10.582\pm0.884$ & & & & \\
      \bottomrule
    \end{tabular}}
  \end{table*}

  DeepPySR achieves the best $R^2$ at every age from 10 onward ($0.604\pm0.012$--$0.425\pm0.012$), outperforming Random Forest (best conventional model at age 8, $R^2=0.765$) at all later ages.  The advantage over PySR grows with age: PySR degrades sharply ($R^2=0.665\to0.210$) as dimensionality overwhelms its search; DeepPySR retains a $0.215$ advantage at age 27.  Per-age Wilcoxon tests confirm DeepPySR significantly outperforms MLP and KAN at all seven ages ($p=0.031$), XGBoost at six of seven, and ElasticNet from age 20 onward; a longitudinal test pooling mean R$^2$ across ages confirms the cumulative advantage over PySR is significant ($p=0.031$).  KAN completely fails ($R^2=0.000$ at all ages).  The interpretable formula ($R^2=0.674$--$0.354$ across ages 8--27) consistently outperforms PySR and all conventional baselines at ages 10--27.  DVPS/EPS ablation results, Pareto-front visualisations, and per-age circle plots are in Supplementary Note~S3.

  This benchmark illustrates DVPS's core advantage at its most extreme.  The 125 features include near-collinear anthropometric measures that confound standard evolutionary search.  More critically, standard feature selection universally discards two key perinatal variables: maternal asthma ($\mathrm{asthma}_m$, 19\% prevalence) and paternal bereavement ($\mathrm{death}_p$, 2\% prevalence) — recursive feature elimination and impurity-based importance assign both negligible scores.  DVPS instead retains a feature only if it reduces formula loss on held-out data, allowing sparse but informative variables to survive regardless of marginal prevalence.  The discovered formula retains both perinatal adversity variables despite their rarity, surfacing an interaction that no standard ML pipeline would have preserved.

  The \textbf{interpretable formula} (29 nodes; $R^2 = 0.674$--$0.354$ across ages 8--27), selected by EPS with $r^2$-weight~$= 1.0$ and $\lambda = 0.005$:
  \begin{align}
    \hat{y} &= z_5^{\mathrm{BMI}} + \Bigl(\mathrm{PRS}_{\mathrm{BMI}} +
    \frac{0.74\,e^{\,\phi}}{\mathrm{mht\_wt}}
    + 8.37\Bigr)\log(\mathrm{age}), 8\le age\le27 \label{eq:bmi_full} \\
    \phi &= z_5^{\mathrm{BMI}} + \bigl(\mathrm{PRS}_{\mathrm{BMI}} + z_1^{\mathrm{BMI}}\bigr)
    (-\mathrm{asthma}_m + \mathrm{death}_p), \nonumber
  \end{align}
  where $\mathrm{PRS}_{\mathrm{BMI}}$ is the polygenic risk score (PGS002313), $z_5^{\mathrm{BMI}}$ and $z_1^{\mathrm{BMI}}$ are BMI z-scores at ages 5 and 1, $\mathrm{mht\_wt}$ is maternal height-weight ratio, $\mathrm{asthma}_m$ is maternal asthma, and $\mathrm{death}_p$ is paternal bereavement.  At a higher conceptual level, the dominant structure of Eq.~\ref{eq:bmi_full} can be written as:
  \begin{equation}
    \hat{y} \approx z_5^{\mathrm{BMI}} +
    \mathrm{PRS}_{\mathrm{BMI}} \cdot e^{\,z_5^{\mathrm{BMI}}} \cdot \log(\mathrm{age}), 8\le age\le27
    \label{eq:bmi_concept}
  \end{equation}
  capturing that adult BMI is governed by the interaction of genetic risk ($\mathrm{PRS}_{\mathrm{BMI}}$) with early childhood growth ($e^{z_5^{\mathrm{BMI}}}$), scaled logarithmically with prediction age.  The full formula (Eq.~\ref{eq:bmi_full}) extends this with a perinatal modulation term $\phi$ (adversity-weighted interaction of PRS and early BMI z-score) and a maternal health denominator ($\mathrm{mht\_wt}$), achieving R$^2 = 0.674$--$0.354$ across ages 8--27~\cite{warrington2019maternal}.

  \textbf{Biological interpretation.}  Each term in Eq.~\ref{eq:bmi_full} corresponds to a distinct biological mechanism.  The dominant structure captures the gene–environment interaction hypothesis of adiposity: PRS$_{\mathrm{BMI}}$ \emph{amplifies} the early-childhood trajectory ($z_5^{\mathrm{BMI}}$) logarithmically with age, consistent with genetic effects on BMI becoming more pronounced over adolescence~\cite{warrington2019maternal}.  The perinatal adversity term $\phi$ encodes an early-life stress pathway: parental health adversity has been linked to altered HPA-axis programming and elevated offspring adiposity~\cite{entringer2012prenatal,felix2019cohort}.  The maternal height-weight denominator ($\mathrm{mht\_wt}$) captures intrauterine growth constraint: lower $\mathrm{mht\_wt}$ \emph{increases} predicted BMI, consistent with the fetal origins hypothesis~\cite{barker1995fetal}.  Together, the formula integrates three biologically separable pathways — genetic programming, early developmental trajectory, and perinatal adversity — without any domain knowledge supplied to the search.

  Figure~\ref{fig:convergence}e shows the DeepPySR relationship circle plot for Raine BMI. Strong Layer-1 arcs connect BMI to PRS and early childhood BMI z-score, while sociodemographic variables appear in outer layers, directly visualising how the multi-layer architecture decomposes ICC clusters.

  \section{Discussion}
  \label{sec:discussion}

  DeepPySR demonstrates that evolutionary symbolic regression, augmented with principled feature pruning and formula selection, can match or exceed conventional machine learning across balanced regression, noisy regression, binary and multiclass classification under severe imbalance, and high-dimensional longitudinal prediction with ICC, while returning human-readable equations.  A discovered symbolic formula is directly inspectable, validatable against domain theory, and communicates through mechanism rather than correlation — properties fundamentally unachievable through post-hoc explanation of a black-box model~\cite{rudin2019stop}.

  The dynamic variable pruning schedule addresses the primary practical barrier to applying SR beyond toy benchmarks: the curse of dimensionality.  By embedding adaptive feature elimination within the evolutionary loop rather than as a static pre-processing step, DVPS allows search to self-organise around causal variables as iterations progress.  Across 6 of 7 real-world datasets, DVPS configurations outperform the no-pruning PySR baseline; the benefit scales with dimensionality, from negligible for four-variable physics equations to a $13\times$ reduction in MSE on the 125-feature Raine BMI task.  Crucially, evolutionary fitness operates on held-out formula accuracy rather than marginal predictor importance, so DVPS retains sparse but informative features such as \texttt{death\_p} (2\% prevalence) and \texttt{asthma\_m} (19\%) that impurity-based selection universally discards.

  The exponential Pareto selection criterion resolves the formula-selection problem that is otherwise left to heuristics or manual inspection.  By applying an exponential penalty to complexity with a user-tunable $\lambda$ and $r^2$-weight $\rho$, EPS searches the existing hall-of-fame for the formula with the best accuracy–complexity trade-off under a coherent scoring function. In practice, EPS reliably identifies formulas that retain 97--100\% of peak performance at 30--50\% lower node count than naive accuracy maximisation; Pareto-front visualisations across all datasets (Supplementary Note~S3, Fig.~S3) confirm that EPS consistently selects formulas at or near the convex knee, where RMSE improvement saturates.  Both hyperparameters have intuitive semantics: a practitioner who needs a formula that is communicable to a non-technical audience can increase $\lambda$ (to steepen the complexity penalty) or decrease $\rho$ (downweight accuracy); one who prioritises predictive performance can decrease $\lambda$ or increase $\rho$.  This makes formula selection reproducible and non-arbitrary, addressing a longstanding gap between SR's Pareto-front output and end-user formula deployment.

  The hierarchical SR architecture resolves ICC by construction: the outer layer identifies a single representative from each correlated cluster, and inner layers model its functional relationship with cluster members.  Regarding student performance, G2 serves as the Layer-1 anchor for G3 prediction, with G1 appearing in the inner layers as a modulator, consistent with the educational research finding that trajectory (not just current level) predicts final performance. In Raine BMI, the outer layer selects early BMI z-score as the dominant anchor, while PRS and sociodemographic variables enter as inner-layer modulators.  The relationship circle plots (Fig.~\ref{fig:convergence}d--e) make this structure visually explicit, providing an interpretable summary of the ICC decomposition that is not available from any conventional model.

  KAN performs comparably to conventional baselines on clean low-dimensional data, but collapses on classification tasks (F1 below random for stroke and diabetes) and on any dataset with categorical inputs or high ICC.  A recent study on longitudinal BMI prediction~\cite{chen2026longitudinal} demonstrates that KAN succeeds, but only on a curated 20- 30-feature input that required substantial domain-guided feature selection prior to modelling. This prerequisite is unavailable in truly exploratory settings, and our Raine BMI task (125 raw features, no pre-selection) is specifically designed to stress-test. These results confirm that KAN does not provide a practical alternative to evolutionary SR when the relevant feature set is unknown in advance.

  The heart disease formula is fully auditable: each variable has a documented clinical rationale, and it is short enough to fit on a reference card.  The stroke formula recovers over half of all true stroke cases without the accuracy collapse observed in SMOTE-based methods.  The Raine BMI formula encodes a biologically plausible interaction between genetic susceptibility and developmental trajectory~\cite{warrington2019maternal} — a functional form that no post-hoc attribution method could have produced, because discovering it requires searching rather than ranking.

  SR remains computationally expensive: the 27-configuration grid with 5-fold cross-validation requires 70-150 CPU-hours per dataset.  Recovered formulas depend on the operator set; Feynman I.9.18 (gravitational force) was not exactly recovered because its division-by-sum-of-squared-differences structure is difficult to evolve without structural guidance.  Classification under extreme imbalance remains challenging, and the multi-layer architecture uses a fixed number of layers.  Future work will investigate uncertainty quantification via ensemble SR, dimensional analysis constraints~\cite{cranmer2023interpretable}, and automated depth selection for the multi-layer architecture.

  \section{Methods}
  \label{sec:methods}

  \subsection{Symbolic Regression Preliminaries}
  \label{sec:sr_prelim}

  Let $\mathcal{D} = \{(\mathbf{x}_i, y_i)\}_{i=1}^N$ be a dataset with $\mathbf{x}_i \in \mathbb{R}^d$ and $y_i \in \mathbb{R}$.  SR seeks an expression tree $f$ drawn from a grammar $\mathcal{G}$ of operators (e.g., $+, \times, \sin, \exp$) and operands (input variables or constants) such that $f(\mathbf{x}) \approx y$.  The search space is combinatorially large, so evolutionary algorithms evolve a population of candidate trees by repeated mutation and selection.

  We build on deeppysr.jl (the Julia back-end of PySR)~\cite{cranmer2023interpretable}, which runs multiple independent evolutionary populations in parallel and periodically migrates the best individuals across populations.  The cost of each expression combines predictive loss $\mathcal{L}$ and a parsimony penalty proportional to the number of nodes~$c$:
  \begin{equation}
    \mathrm{cost}(f) = \mathcal{L}(f) + \beta \cdot c(f),
    \label{eq:base_cost}
  \end{equation}
  where $\beta$ is the parsimony coefficient.  After the search, PySR returns a Pareto front of non-dominated (loss, complexity) pairs.

  \subsection{Dynamic Variable Pruning Schedule (DVPS)}
  \label{sec:dvps}

  In high-dimensional datasets ($d \gg 1$), expression trees tend to include many irrelevant variables early in the search because any variable can reduce loss slightly by overfitting.  Static parsimony penalises tree size but does not specifically target variable diversity.  We introduce a dedicated mutation operator, \texttt{prune\_variable}, that replaces a leaf variable node with a numeric constant.  Its weight in the mutation schedule grows linearly as the search matures.

  At iteration $t$, the effective weight of the pruning mutation is:
  \begin{equation}
    w_{\mathrm{prune}}(t) = w_0 \cdot \tau(t) \cdot M,
    \label{eq:prune_schedule}
  \end{equation}
  \begin{equation}
    \tau(t) = \min\!\left(
                   1,\, \max\!\left(
                       0,\,
                       \frac{t - T_{\mathrm{start}}}{T_{\mathrm{ramp}}}
                    \right)
                \right),
  \end{equation}
  where $w_0$ is the base pruning weight (default $1.0$), $T_{\mathrm{start}}$ is the iteration at which pruning begins, $T_{\mathrm{ramp}}$ is the ramp duration over which the weight linearly increases from $0$ to full, and $M$ is the maximum multiplier cap.  Before $T_{\mathrm{start}}$, $\tau(t) = 0$ and no pruning occurs; after $T_{\mathrm{start}} + T_{\mathrm{ramp}}$, $\tau(t) = 1$ and pruning is applied at full weight $w_0 \cdot M$.

  This schedule has three desirable properties: (i) early iterations are unaffected, allowing exploration of all features; (ii) pruning pressure ramps smoothly, preventing sudden disruption of the population; (iii) a cap $M$ prevents pruning from overwhelming other mutation operators.  The schedule is implemented in \texttt{Mutate.jl} of deeppysr.jl.

  Algorithm~\ref{alg:dvps} summarises the full evolutionary loop with DVPS.

  \begin{algorithm}
    \caption{Evolutionary SR with Dynamic Variable Pruning Schedule}
    \label{alg:dvps}
    \begin{algorithmic}[1]
      \State \textbf{Input:} dataset $\mathcal{D}$, operators $\mathcal{G}$,
      params $T_{\mathrm{start}}, T_{\mathrm{ramp}}, M, w_0$
      \State Initialise population $\mathcal{P}$ of random expression trees
      \For{$t = 1, \ldots, T_{\mathrm{max}}$}
        \State $\tau \gets \min(1, \max(0, (t - T_{\mathrm{start}})/T_{\mathrm{ramp}}))$
        \State $w_{\mathrm{prune}} \gets w_0 \cdot \tau \cdot M$
        \State Update mutation weights with $w_{\mathrm{prune}}$
        \ForAll{individuals $f \in \mathcal{P}$}
          \State Sample mutation type from updated weight distribution
          \If{mutation = \texttt{prune\_variable}}
            \State Replace a random variable node with a constant
          \Else
            \State Apply standard mutation (e.g., \texttt{mutate\_operator})
          \EndIf
          \State Evaluate $\mathrm{cost}(f)$ on $\mathcal{D}$
          \State Update Pareto front
        \EndFor
        \State Migrate top individuals across sub-populations
      \EndFor
      \State \textbf{Return:} Pareto front of (loss, complexity) pairs
    \end{algorithmic}
  \end{algorithm}

  \subsection{Exponential Pareto Selection (EPS)}
  \label{sec:eps}

  After evolutionary search, SR produces a Pareto front $\mathcal{F} = \{f_1, f_2, \ldots\}$ of non-dominated solutions indexed by accuracy and complexity.  Selecting the best single formula is non-trivial: purely accuracy-based selection favours the most complex equation; purely complexity-based selection discards high-performing models.  We define the following scoring function:
  \begin{equation}
    \mathrm{score}(f_i) =
    \bigl[\max(10^{-4},\, R^2_i)\bigr]^{\rho}
    \cdot \exp\bigl(-\lambda\,(c_i - 1)\bigr),
    \label{eq:eps}
  \end{equation}
  where $R^2_i$ is the cross-validated R$^2$ of equation $f_i$, $c_i$ is its node count, $\rho$ is a user-tunable accuracy exponent ($\rho \geq 1$), and $\lambda > 0$ is a user-tunable complexity decay constant.  The selected formula is $f^* = \arg\max_i \mathrm{score}(f_i)$.

  The first factor rewards accuracy superlinearly (when $\rho > 1$), magnifying differences between good and mediocre formulas.  The exponential decay penalises complexity multiplicatively: each additional node reduces the score by $e^{-\lambda}$, preventing complex equations from surviving with only marginally higher R$^2$.  Both $\rho$ and $\lambda$ are user-tunable and grid-searched via cross-validated performance.  Crucially, this grid search is computationally negligible. It scores equations already stored in the evolutionary hall of fame rather than re-running the model, so the entire sweep completes in seconds. Users can supply domain-specific candidate pairs: a clinical audit requiring compact formulas would set a high $\lambda$, while an automated prediction task might set a low $\lambda$ and a high $\rho$.  The floor $\max(10^{-4}, R^2_i)$ prevents numerical issues when $R^2 \leq 0$.

  \subsection{Multi-Layer Symbolic Regression}
  \label{sec:deepsr}

  Many real-world systems exhibit hierarchical functional structure, and in the presence of ICC, standard SR may converge to arbitrary combinations of mutually predictive variables.  The multi-layer architecture addresses this by treating correlated predictor clusters as subtrees: the outer function selects one representative compound feature, and inner layers reveal how that feature is constituted by correlated predictors.

  The architecture discovers:
  \begin{equation}
    y = f\bigl(g_1(\mathbf{x}),\, g_2(\mathbf{x}),\, \ldots,\, g_K(\mathbf{x})\bigr),
    \label{eq:deep_sr}
  \end{equation}
  where $f$ and each $g_k$ are symbolic expressions, and $K$ is the number of latent variables.  Equation~\ref{eq:deep_sr} illustrates the two-layer case, the primary architecture evaluated here; the recursion naturally generalises to additional depths.  The algorithm uses a breadth-first queue: (1) run SR on the original target $y$ against raw features to discover the outer expression $\hat{f}$; (2) enqueue each free variable $x_k$ appearing in $\hat{f}$ as a latent regression target for the next layer; (3) for each dequeued target $x_k$, run SR using raw features that exclude $x_k$ itself and its direct parent variable in the hierarchy; (4) substitute the discovered sub-expression $\hat{g}_k$ symbolically back into $\hat{f}$; (5) continue until the queue is empty or the maximum layer depth is reached.

  \textbf{Stopping criterion.}  Each latent target $x_k$ is assigned an EPS score after its sub-expression is discovered.  If this score falls below a user-specified threshold (default 2.0), $x_k$ is declared a leaf node: its raw value is retained in the composed formula without further decomposition.

  \textbf{Circular dependency prevention.}  When fitting a latent target $x_k$, only $x_k$ itself and its direct parent variable are removed from the feature matrix.  All other variables, including grandparents, siblings, and variables at deeper layers, remain available as candidate inputs.  This deliberately local exclusion prevents immediate self- and parent-references while preserving the search's ability to discover ICC relationships across the full input space.  For example, if the outer formula links BMI to a PGS score and $z_5^{\mathrm{BMI}}$, the SR run for the PGS score excludes only the PGS column itself and the BMI target column, but retains $z_5^{\mathrm{BMI}}$ and all other variables as candidate features.

  \textbf{Error propagation.}  The final prediction is obtained by symbolic substitution of all $\hat{g}_k$ into $\hat{f}$ using SymPy, followed by \texttt{lambdify} for vectorised evaluation.  Errors at each layer are carried forward algebraically through the composition rather than accumulated numerically.

  Each SR call uses DVPS and EPS.  Runtime scales with the number of input variables and model depth, as each layer requires an independent SR run.

  \subsection{Datasets}
  \label{sec:datasets}

  \textbf{Feynman physics benchmarks}~\cite{udrescu2020ai}.  Four equations: Gaussian (I.6.2a; 1 var), kinetic energy (I.13.4; 4), gravitational force (I.9.18; 7), electromagnetic scattering (I.32.17; 6).  $N = 1{,}000$ noise-free samples each.  Full metadata in Supplementary Note~S1.1.

  \textbf{Body fat}~\cite{johnson1996fitting} (252 samples, 14 features; regression).  Body fat percentage with 13 highly intercorrelated anthropometric measurements (high ICC).  Critical challenge: multicollinearity among Weight, Abdomen, and other circumferences.  Full description in Supplementary Note~S1.2.

  \textbf{Heart disease}~\cite{detrano1989international} (303 samples, 13 features; binary classification).  Cleveland Heart Disease; mixed clinical measurements; moderate class imbalance (54\% positive).  Full description in Supplementary Note~S1.3.

  \textbf{Wine quality}~\cite{cortez2009modeling} (Red: 1,599; White: 4,898; 11 features; regression).  Physicochemical measurements with noisy subjective quality ratings; red and white modelled separately.  Critical challenge: label noise and correlated physicochemical features.  Full description in Supplementary Note~S1.4.

  \textbf{Stroke} (5,110 samples, 10 features; binary classification).  Medical history and demographics; severe class imbalance (4.9\% positive stroke). Critical challenge: extreme minority-class recovery without SMOTE.  Full description in Supplementary Note~S1.5.

  \textbf{Student performance}~\cite{cortez2008student} (Math: 395; Portuguese: 649; 33 features; regression).  Demographic, family, and school attributes.  Critical challenge: strong ICC among grade variables (G1, G2, G3); many categorical features.  Full description in Supplementary Note~S1.6.

  \textbf{Raine Generation 2 BMI longitudinal}~\cite{straker2017raine}.  Prospective birth cohort (Western Australia).  Task: predict BMI at ages 8, 10, 14, 17, 20, 23, 27 from early childhood measurements (birth, year 1, year 5 only). 125 candidate features including anthropometric measures, polygenic risk scores (PGS002313, PGS002853, PGS002161), and sociodemographic variables. Critical challenges: high ICC, severe missingness, longitudinal targets. Full metadata and PRS provenance are provided in Supplementary Notes~S1.7 and Supplementary Data~1--2; the complete data preprocessing pipeline is illustrated in Supplementary Note~S1.7.5, Fig.~S1.

  \textbf{Diabetes}~\cite{cdc2015brfss} (253,680 samples, 21 features; multiclass: no diabetes / prediabetes / diabetes; class distribution 84.2\% / 1.8\% / 13.9\%).  CDC BRFSS self-reported health indicators. Critical challenge: severe class imbalance with prediabetes class being heavily under-represented.  Full description in Supplementary Note~S1.8.

  No feature engineering is applied; all models receive raw features (label-encoded or standardised).

  \subsection{Experimental Setup}
  \label{sec:setup}

  All experiments use 5-fold cross-validation with stratified splits for classification tasks and covariate-stratified splits for regression.  A held-out test set was not used because all public datasets were used in full; no external dataset of the same structure was available for independent validation, and a fixed split on small datasets would produce high-variance estimates.  The Raine cohort is the sole private dataset and cannot be shared externally for out-of-sample evaluation.  DeepPySR and PySR receive raw features with \emph{no} preprocessing: no feature selection, no normalisation, and no SMOTE oversampling.  For class-imbalanced datasets, SMOTE is applied only to baseline classifiers (XGBoost, Random Forest, Extra Trees, MLP, ElasticNet) to ensure a fair comparison; SR models are deliberately evaluated without it to assess whether the formula structure alone can handle the imbalance.

  The use of SMOTE for baselines but not for symbolic regression models was a deliberate design choice motivated by the extreme class imbalance present in three datasets: diabetes (prediabetes class $< 2\%$), stroke ($\approx 4.9\%$ positive), and heart disease (54\% positive but with clinically meaningful minority-class structure).  Without SMOTE, baseline classifiers collapsed toward majority-class prediction, producing F1 scores near zero for minority classes and rendering the comparison uninformative.  SMOTE was therefore applied to baselines to establish competitive, meaningful benchmarks.  Even with this advantage, baseline F1 scores on stroke and diabetes remain substantially below those achieved by DeepPySR and PySR.  Symbolic regression models were deliberately not given SMOTE because evaluating their intrinsic formula structure under real-world imbalance conditions is itself a research question: the goal was to test whether a compact algebraic formula can encode the class boundary without synthetic oversampling.  This asymmetric treatment favours baselines, thereby making the SR results conservative.

  We grid-search 27 SR configurations: pruning parameters ($T_{\mathrm{start}} \in \{25, 50, 75\}$, $T_{\mathrm{ramp}} \in \{50, 100, 150\}$) and adaptive parsimony scaling ($\alpha \in \{1.0, 10.0, 50.0\}$). EPS hyperparameters ($\rho \in \{1.0, 1.5, 2.0\}$, $\lambda \in \{0.001, 0.005, 0.01\}$) are searched independently over the resulting Pareto fronts. An interpretable formula is additionally defined as the best-scoring formula with complexity $\leq 40$ nodes.  Following cross-validated training, we conduct (a) multi-layer analysis on the full dataset with a circle plot visualisation; (b) Wilcoxon signed-rank tests comparing DeepPySR and PySR performance distributions across folds; and (c) convergence analysis tracking per-iteration R$^2$ or F1.

  All SR runs use: binary operators $\{+, -, \times, \div, \mathrm{piecewise}\}$ and unary operators $\{\exp, \log, \sin, \sqrt{\,\cdot\,}\}$; maximum tree size 40 per SR call (where \emph{tree size} is the total node count: every operator, variable, and constant in the expression tree; this hard limit is enforced during the evolutionary search so that the hall-of-fame stores only trees with at most 40 nodes; in the multi-layer architecture, sub-expressions from inner layers are substituted into the outer formula symbolically after search, so the final composed formula can substantially exceed this limit); 100 populations of 200 individuals; 200 evolution cycles per iteration; 500 iterations (100 for stroke and diabetes); parsimony coefficient $\beta = 0.001$; pruning cap $M = 0.7$.

  Baselines: XGBoost~\cite{chen2016xgboost}, Random Forest~\cite{breiman2001random}, Extra Trees~\cite{geurts2006extremely}, ElasticNet/LogisticRegression, MLP (two hidden layers, 128–64 units, ReLU, Adam), KAN~\cite{liu2024kan}, and PySR (vanilla, without DVPS, same operator set and iteration budget).

  For regression: R$^2$, RMSE (MAE in Supplementary Note~S2).  For classification: accuracy, precision, recall, F1.  Formula complexity is the number of nodes in the expression tree.

  \subsection{Computational Resources}
  \label{sec:compute}

  Experiments were conducted on the Setonix supercomputer at the Pawsey Supercomputing Research Centre and the ICRAR computing cluster.  Each SR configuration was run as an independent parallel job.  The 27-configuration grid with 5-fold cross-validation requires approximately 70-150 CPU-hours per dataset.

  \section*{Ethics statement}

  This study, utilising de-identified data from the Raine Study Gen2 cohort, received an exemption from ethics review by the University of Western Australia's Human Research Ethics Office (Ref: 2025/ET000396, April 28, 2025), as no new data from human subjects were collected. Informed consent was obtained from participants' guardians during the original Raine Study, in accordance with institutional protocols. Data privacy and confidentiality were ensured through de-identification and secure storage, compliant with Australian data protection guidelines. No compensation was provided to participants, as the study relied on existing data.

  \section*{Data Availability}

  The Feynman equations benchmark is publicly available at \url{https://space.mit.edu/home/tegmark/aifeynman.html}. Body fat, heart disease, wine quality, stroke, and student performance datasets are available from the UCI Machine Learning Repository. The diabetes dataset is available from the CDC BRFSS repository. Raine Generation 2 data are available to approved researchers through application to the Raine study management committee.

  \section*{Code Availability}

  DeepPySR is implemented as \texttt{deeppysr.jl} (Julia back-end, \url{https://github.com/ICRAR/deeppysr.jl}) with a Python wrapper package (\url{https://github.com/ICRAR/DeepPySR}); both are publicly available on GitHub.  All experiment scripts reported in this paper are available in the \texttt{DeepPySR} repository.

  \section*{Acknowledgements}

  We gratefully acknowledge all Raine Study participants and their families for their continued participation in the study, as well as the Raine Study team for study coordination and data collection. We also thank the NHMRC and the Raine Medical Research Foundation for their support. The core management of the Raine Study is funded by The University of Western Australia, Curtin University, The Kids Research Institute Australia, Women and Infants Research Foundation, Edith Cowan University, Murdoch University, The University of Notre Dame Australia and the Western Australian Future Health Research and Innovation Fund (2023--2024; Grant ID WACSOSP2023-2024). The Pawsey Supercomputing Centre (Setonix) provided computational resources to carry out analyses required with funding from the Australian Government and the Government of Western Australia. The data collection of the Raine Study Gen1- and 2-1, 2, 5, 8, 10, 14, 17, 20, 22, and 26 year follow-ups were funded by NHMRC project grants (211912, 003209, 572613, 403981, 353514, 572613, 403981, 1059711, 1084947), and The Raine Medical Research Foundation.

  \section*{Author contributions}

  F.C.\ designed and implemented DeepPySR, performed all experiments, and wrote the manuscript.  K.V.\ contributed to the evaluation and analysis of results. P.M.\ contributed to the evaluation and analysis of the Raine Generation 2 cohort. R.-C.H.\ contributed to the evaluation and analysis of the Raine Generation 2 cohort.  All authors reviewed and approved the final manuscript.

  \section*{Competing interests}

  The authors declare no competing interests.

  \bibliographystyle{unsrtnat}
  \bibliography{refs}

\end{document}